\documentclass[10pt,twocolumn]{article}

\usepackage[utf8]{inputenc}
\usepackage[T1]{fontenc}
\usepackage{times}
\usepackage{microtype}
\usepackage{graphicx}
\usepackage{booktabs}
\usepackage{amsmath,amssymb}
\usepackage{hyperref}
\usepackage{xcolor}
\usepackage[numbers,sort&compress]{natbib}
\usepackage{caption}
\usepackage{tabularx}
\usepackage{float}
\usepackage{authblk}
\usepackage{tikz}
\usetikzlibrary{positioning,arrows.meta,calc}
\usepackage[margin=1in]{geometry}

\hypersetup{
  colorlinks=true,
  linkcolor=blue!60!black,
  citecolor=blue!60!black,
  urlcolor=blue!60!black,
}

\title{\Large \bf TotalSegmentator Aorta Report: Automated Aortic Diameter Measurement at 11 Standardized Landmarks on CT Images
}

\author[1]{Jakob Wasserthal\thanks{Corresponding author: \texttt{jakob.wasserthal@usb.ch}}}
\author[1]{Daniel Hinck}
\author[1]{Martin Segeroth}
\author[1]{Joshy Cyriac}
\author[2]{Denis Berdajs}
\author[2]{Jules Miazza}
\author[1]{Maurice Pradella}

\affil[1]{Clinic of Radiology and Nuclear Medicine, University Hospital Basel, Basel, Switzerland}
\affil[2]{Department of Cardiac Surgery, University Hospital Basel, University of Basel, Basel, Switzerland}
\date{}

\begin{document}
\maketitle
\thispagestyle{empty}

\begin{abstract}
\textbf{Purpose:} To develop and evaluate a fully automated, open-source tool that measures aortic diameters at 11 standardized landmarks from the aortic annulus to the bifurcation on any CT image, following current guideline recommendations.

\textbf{Materials and Methods:} This retrospective single-center study included 239 arterial phase CT angiography examinations of 230 patients (median age, 65 years [range, 18--99 years]; 163 men) performed between April 2025 and October 2026 (209 electrocardiographically [ECG] gated). The tool locates landmarks from TotalSegmentator segmentations of aortic branches and vertebrae and measures the maximum diameter at each. A senior radiologist reviewed and, where needed, corrected the automated measurements to create the reference standard. Agreement was assessed with mean absolute error (MAE), bias, and differences greater than 2\,mm.

\textbf{Results:} Across all landmarks, MAE was 1.06\,mm (95\% CI: 0.91, 1.22), and 86.6\% (95\% CI: 84.4, 88.5) of differences were 2\,mm or less. The tool measured slightly larger diameters (bias, 0.45\,mm [95\% CI: 0.34, 0.57]). The sinuses of Valsalva had the largest MAE (2.17\,mm [95\% CI: 1.59, 2.86]). For 1.2\% of landmarks, the automatic measurement failed. Examination-level MAE did not differ significantly by pathology, atherosclerosis grade ($P=.95$), or ECG gating ($P=.14$). Examinations with mural thrombus more often had an MAE greater than 2\,mm (50\% [95\% CI: 31, 69] vs 20\%; $P=.03$).

\textbf{Conclusion:} The tool measured aortic diameters at 11 landmarks along the entire aorta with an MAE of about 1\,mm, below published interreader variability. Agreement was lower at the aortic root and with mural thrombus.

\textbf{Code availability:} The tool is open source and freely available at \url{https://github.com/wasserth/TotalSegmentator}.
\end{abstract}

\section{Introduction}
\label{sec:introduction}

Aortic diameter is the main measurement for the diagnosis, surveillance, and treatment planning of aortic disease. Guidelines recommend reporting diameters at defined anatomic landmarks, measured perpendicular to the vessel axis, and base the timing of follow-up and intervention on these values~\citep{hiratzka2010,erbel2014,isselbacher2022}. Manual measurement in double-oblique reformations is time-consuming, and its reproducibility is limited: differences between readers of 1.6--5\,mm are common, which is in the same range as the growth that guidelines regard as clinically relevant~\citep{elefteriades2020,goldstein2015}. Aortic dilatation is also underreported when diameters are not measured systematically~\citep{pradella2022}. Automated measurement could make reporting faster and more consistent~\citep{rueckel2021}.

Several deep learning tools measure aortic diameters at CT. The most widely studied commercial tool measures nine thoracic landmarks and agrees with readers to within 1.3--2.6\,mm on average~\citep{pradella2021,monti2022,hamelink2023,graby2023}. However, its validations were mostly restricted to one scan type, such as ECG-gated CT angiography (CTA)~\citep{pradella2021} or low-dose unenhanced CT~\citep{hamelink2023}, it does not cover the abdominal aorta below the celiac trunk, and its errors were largest at the aortic root~\citep{graby2023,lopiccolo2023}. Other tools report only the maximum diameter of a thoracic segment~\citep{macruz2022}, the maximum diameter per segment of the whole aorta~\citep{paulsen2026,postiglione2024}, or a single maximum diameter of the abdominal aorta~\citep{rao2026}. Openly available tools that measure diameters at standardized landmarks along the entire aorta, from the annulus to the bifurcation, remain scarce. Such landmark-based measurements are needed for structured reporting and for comparing the same location at follow-up.

TotalSegmentator is a widely used open-source model for segmenting anatomic structures at CT~\citep{wasserthal2023}. We extended it with an aorta report that places 11 landmarks along the aortic centerline and measures the diameter at each. The tool is designed to work on any CT image, independent of contrast phase and ECG gating. The purpose of this study was to develop this tool and to evaluate its agreement with radiologist-corrected measurements at arterial phase CTA, overall, per landmark, and in subgroups defined by aortic pathology, atherosclerosis, and ECG gating.

\section{Materials and Methods}
\label{sec:methods}

\subsection{Study Design and Sample}

This retrospective single-center study was approved by the Ethics Committee of Northwest and Central Switzerland (EKNZ, Req-2023-00446), and the requirement for written informed consent was waived.

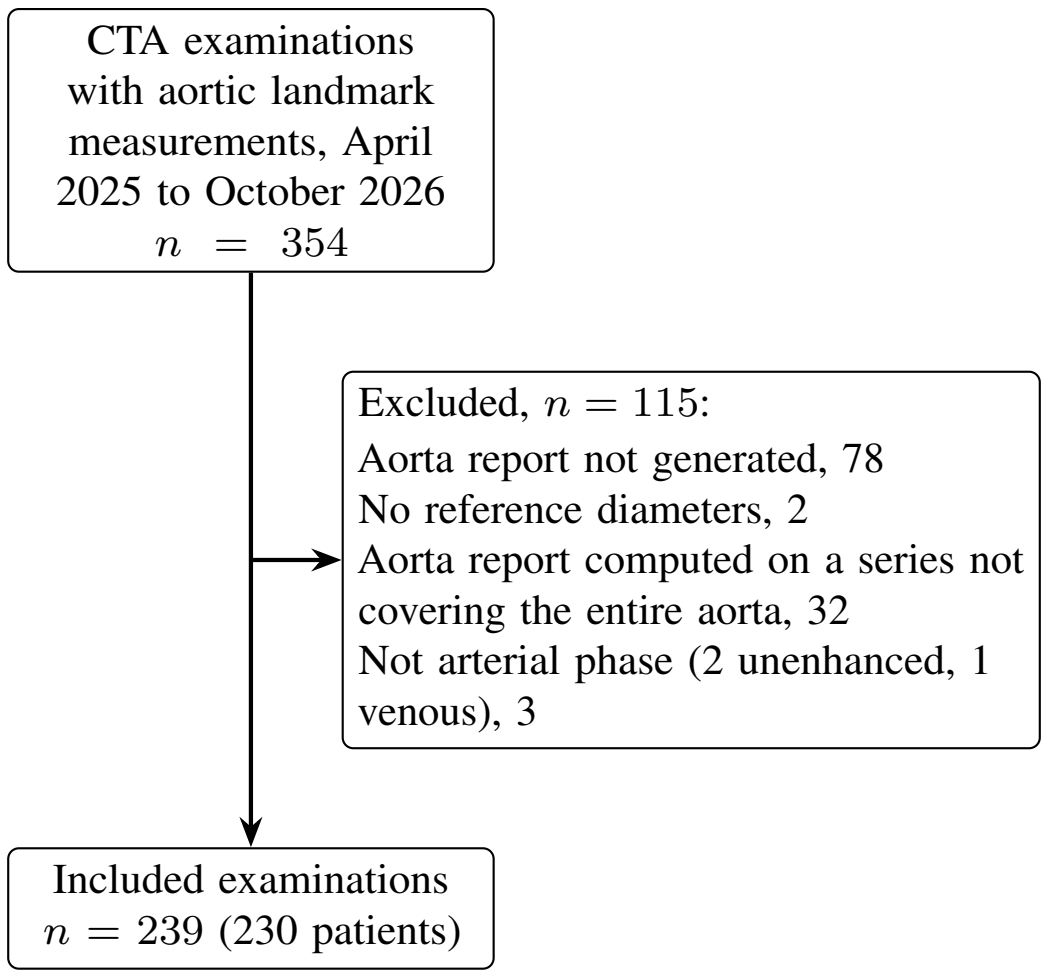
\begin{figure}[t]
  \centering
  \begin{tikzpicture}[
      box/.style={draw, rounded corners=2pt, align=center, font=\footnotesize, text width=30mm, inner sep=3pt},
      exclusion/.style={draw, rounded corners=2pt, align=left, font=\footnotesize, text width=44mm, inner sep=3pt},
      arrow/.style={-{Stealth[length=2mm]}, thick}
    ]
    \node[box] (search) {CTA examinations with aortic landmark measurements, April 2025 to October 2026\\$n=354$};
    \node[box, below=38mm of search] (included) {Included examinations\\$n=239$ (230 patients)};
    \coordinate (mid) at ($(search.south)!0.5!(included.north)$);
    \node[exclusion, anchor=west] (excluded) at ($(mid)+(6mm,0)$) {Excluded, $n=115$:\\[1pt]
      Aorta report not generated, 78\\
      No reference diameters, 2\\
      Aorta report computed on a series not covering the entire aorta, 32\\
      Not arterial phase (2 unenhanced, 1 venous), 3};
    \draw[arrow] (search) -- (included);
    \draw[arrow] (mid) -- (excluded.west);
  \end{tikzpicture}
  \caption{Flowchart of examination selection. CTA = CT angiography.}
  \label{fig:flow}
\end{figure}

CTA examinations with aortic landmark measurements performed between April 2025 and October 2026 were identified by a search of the institutional imaging database with PACScrawler~\citep{cyriac2026}. Exclusion criteria were as follows: the aorta report had not been generated for the examination, no reference diameters were available, the aorta report had been computed on a series that did not cover the entire aorta, or the series was not acquired in the arterial phase (Fig~\ref{fig:flow}). Repeated examinations of the same patient were included as separate examinations. Examination characteristics are summarized in Table~\ref{tab:cohort}.

Contrast phase, ECG gating, and acquisition parameters were read from the DICOM headers and series descriptions.

\begin{table}[t]
  \centering
  \caption{Characteristics of the 239 included CT angiography examinations.}
  \label{tab:cohort}
  \footnotesize
  \begin{tabularx}{\columnwidth}{@{}Xr@{}}
    \toprule
    Characteristic & Value \\
    \midrule
    Age (y)\textsuperscript{a} & 65 (58--74) \\
    Men & 171 (71.5) \\
    Women & 68 (28.5) \\
    ECG-gated, prospective high-pitch & 209 (87.4) \\
    \quad Reconstruction at 65\% / 30\% of R-R interval & 197 / 12 \\
    Not ECG-gated & 30 (12.6) \\
    Scanner: SOMATOM Force / other Siemens & 237 / 2 \\
    Tube voltage (kV)\textsuperscript{a,b} & 100 (90--120) \\
    Section thickness 1.0 / 1.5\,mm\textsuperscript{b} & 207 / 27 \\
    Volume CT dose index (mGy)\textsuperscript{a,b} & 5.8 (3.6--7.8) \\
    \midrule
    Aneurysm & 114 (47.7) \\
    Ectasia or dilatation & 89 (37.2) \\
    Mural thrombus & 22 (9.2) \\
    Aortic graft or stent & 22 (9.2) \\
    Bicuspid aortic valve & 20 (8.4) \\
    Aortic valve prosthesis & 10 (4.2) \\
    Dissection & 9 (3.8) \\
    Atherosclerosis none / mild / moderate / severe\textsuperscript{c} & 76 / 61 / 46 / 19 \\
    \bottomrule
  \end{tabularx}
  \\[2pt]
  \parbox{\columnwidth}{\scriptsize Note.---Unless otherwise indicated, data are numbers of examinations, with percentages in parentheses. The 239 examinations were from 230 patients. ECG = electrocardiography.
  \textsuperscript{a}Data are medians, with IQRs in parentheses.
  \textsuperscript{b}Not available for five examinations.
  \textsuperscript{c}Not graded for 37 examinations.}
\end{table}

\subsection{Automated Aorta Report}

\begin{figure*}[t]
  \centering
  \IfFileExists{imgs/overview.png}{%
    \includegraphics[width=0.85\textwidth]{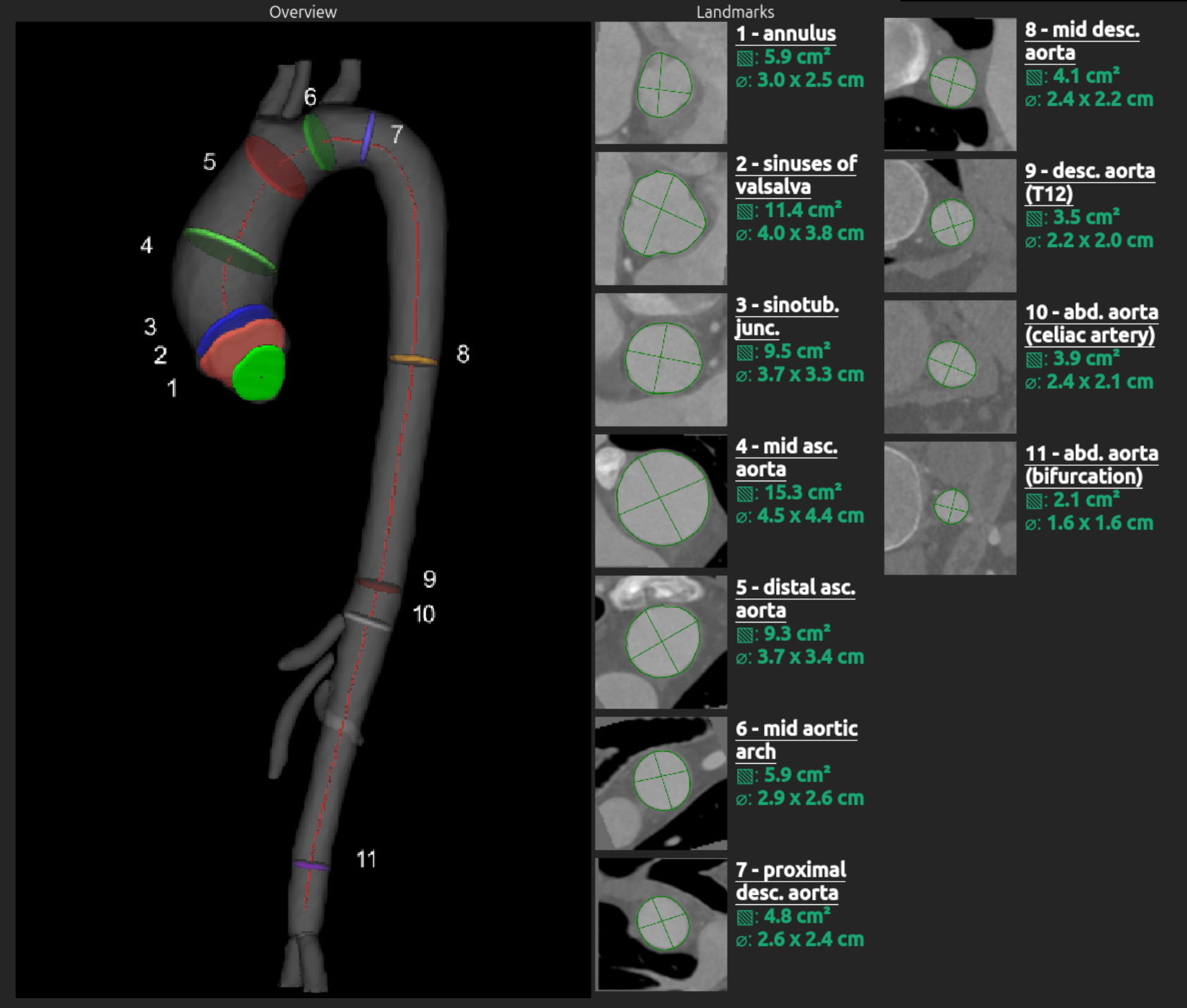}%
  }{%
    \fbox{\parbox{0.85\textwidth}{\centering Missing file: \texttt{imgs/overview.png}}}%
  }
  \caption{Example output of the aorta report in a 46-year-old male patient at ECG-gated arterial phase CT angiography. Left: three-dimensional rendering of the aortic segmentation with the centerline (red dashed line) and the measurement planes at landmarks 1--11. Right: cross-sections perpendicular to the centerline at each landmark, with the cross-sectional area and the maximum and perpendicular diameters (green lines). ECG = electrocardiography.}
  \label{fig:overview}
\end{figure*}

The aorta report is part of TotalSegmentator (version 2.18.0; aorta report version 0.1.1) and runs without user interaction (Fig~\ref{fig:overview}). It uses five segmentation models based on nnU-Net~\citep{isensee2021}. First, a fast TotalSegmentator model segments the aorta and heart, and the CT image is cropped to these structures with a 20-mm margin and resampled to 0.8-mm isotropic voxels. The standard TotalSegmentator model then segments the aorta, the brachiocephalic trunk, the left common carotid and left subclavian arteries, the common iliac arteries, and the T12 vertebra~\citep{wasserthal2023}. Three additional models were trained for the report: one for the aortic annulus and the sinotubular junction (trained on 48 CT examinations), one for the celiac trunk, superior mesenteric artery, and renal arteries (127 examinations), and one for the true and false lumen of aortic dissections (171 examinations). Training images were subsets of the TotalSegmentator dataset~\citep{wasserthal2023}, annotated by a medical expert.

The centerline is extracted by skeletonizing the smoothed aortic mask and keeping the longest path, which is extended to the center of the annulus. The 11 landmarks follow the positions used in structured aortic reporting~\citep{hiratzka2010,isselbacher2022} and are defined relative to the segmented structures (Table~\ref{tab:landmarks}). For each landmark, a plane perpendicular to the local centerline is intersected with the aortic mask. For the annulus, the plane follows the orientation of the annulus mask. To reduce slight oversegmentation, the mask is eroded by one voxel (0.8\,mm) before measurement. The maximum diameter is the longest distance between two points on the boundary of the cross-section. The perpendicular diameter is the longest chord approximately perpendicular to it. A landmark remains empty if a structure it depends on is not found. The report also provides cross-sectional areas, section lengths and volumes, and a curved planar reformation, which were not evaluated here.

\begin{table}[t]
  \centering
  \caption{Landmark definitions of the aorta report.}
  \label{tab:landmarks}
  \footnotesize
  \begin{tabularx}{\columnwidth}{@{}lX@{}}
    \toprule
    Landmark & Position on the centerline \\
    \midrule
    1 Annulus & Center of the annulus segmentation \\
    2 Sinuses of Valsalva & Largest cross-sectional area between landmarks 1 and 3, excluding 10\,mm at both ends \\
    3 Sinotubular junction & Closest to the sinotubular junction segmentation \\
    4 Mid ascending aorta & Midway between landmarks 3 and 5 \\
    5 Distal ascending aorta & 5\,mm proximal to the brachiocephalic trunk origin \\
    6 Mid aortic arch & Midway between landmarks 5 and 7 \\
    7 Proximal descending aorta & 5\,mm distal to the left subclavian artery origin \\
    8 Mid descending aorta & Midway between landmarks 7 and 9 \\
    9 Descending aorta at T12 & Closest to the center of the T12 vertebra \\
    10 Abdominal aorta, celiac trunk & Cranial edge of the celiac trunk origin \\
    11 Abdominal aorta, bifurcation & 20\,mm proximal to the distal end of the centerline \\
    \bottomrule
  \end{tabularx}
\end{table}

\subsection{Reference Standard}

A fellowship-trained cardiothoracic radiologist radiologist (6 years of experience in cardiovascular imaging) reviewed the automated measurements of each examination on the CT images and corrected diameters judged incorrect. Landmarks without an automated measurement were measured manually. The corrected diameters, recorded in whole millimeters, served as the reference standard. Maximum and perpendicular diameters were reviewed in the same way.

\subsection{Aortic Findings}

Aortic findings documented by the reading radiologist were coded by one author (M.P.). These were aneurysm, ectasia or dilatation, mural thrombus (thrombus lining the aortic wall, typically in an aneurysm; intramural hematoma was coded separately), aortic graft or stent, bicuspid aortic valve, aortic valve prosthesis, dissection, and the atherosclerosis grade (none, mild, moderate, or severe). Findings present in fewer than 10 examinations were not analyzed separately.

\subsection{Statistical Analysis}

The difference was defined as the aorta report minus the reference, with positive values indicating larger measurements by the tool. Agreement was summarized by mean absolute error (MAE), bias (mean difference), and 95\% limits of agreement~\citep{bland1986}. Because 63\% of the differences were exactly 0\,mm, the median absolute difference did not discriminate between landmarks or groups, so means are reported. A difference greater than 2\,mm was considered clinically relevant, because this is below the reported interreader variability of CT aortic diameter measurements~\citep{goldstein2015,hamelink2023,singh2003,cayne2004}. A landmark was counted as failed if the reference had a diameter but the tool did not. Landmark 11 was not counted as failed, because the bifurcation is often outside the field of view of thoracic CTA. Its diameter was compared when both values existed.

For each examination, the MAE and bias were averaged over its compared landmarks. Bias was tested against zero with the Wilcoxon signed-rank test. Examinations with and without each finding, and ECG-gated and nongated examinations, were compared with the Mann-Whitney U test for examination-level MAE and bias, and with the Fisher exact test for the proportion of examinations with an MAE greater than 2\,mm and with at least one failed landmark. Atherosclerosis grades were compared with the Kruskal-Wallis test. Maximum and perpendicular diameter errors were compared with the Wilcoxon signed-rank test on paired absolute differences. $P$ values were adjusted with the Holm method across the 11 landmarks or across the six findings~\citep{holm1979}. The 95\% CIs were calculated with the Wilson score method for proportions of examinations and otherwise with a percentile bootstrap (2000 resamples of examinations). Analyses used Python (version 3.11) with SciPy (version 1.14.1). Two-sided $P<.05$ indicated statistical significance. Subgroup and ECG analyses used the maximum diameter.

\section{Results}
\label{sec:results}

\subsection{Study Sample}

Of 354 examinations, 115 were excluded (Fig~\ref{fig:flow}). The 239 included examinations were from 230 patients (163 men and 67 women; median age, 65.5 years [IQR, 58--74 years]; range, 18--99 years). All were arterial phase CTA examinations, 209 of 239 (87.4\%) with prospective high-pitch ECG gating, and 237 of 239 (99.2\%) were acquired on the same scanner model (Table~\ref{tab:cohort}). Aneurysm was documented in 114 examinations and ectasia in 89 examinations.

\subsection{Agreement per Landmark}

\begin{figure*}[t]
  \centering
  \IfFileExists{imgs/landmark_differences.png}{%
    \includegraphics[width=\textwidth]{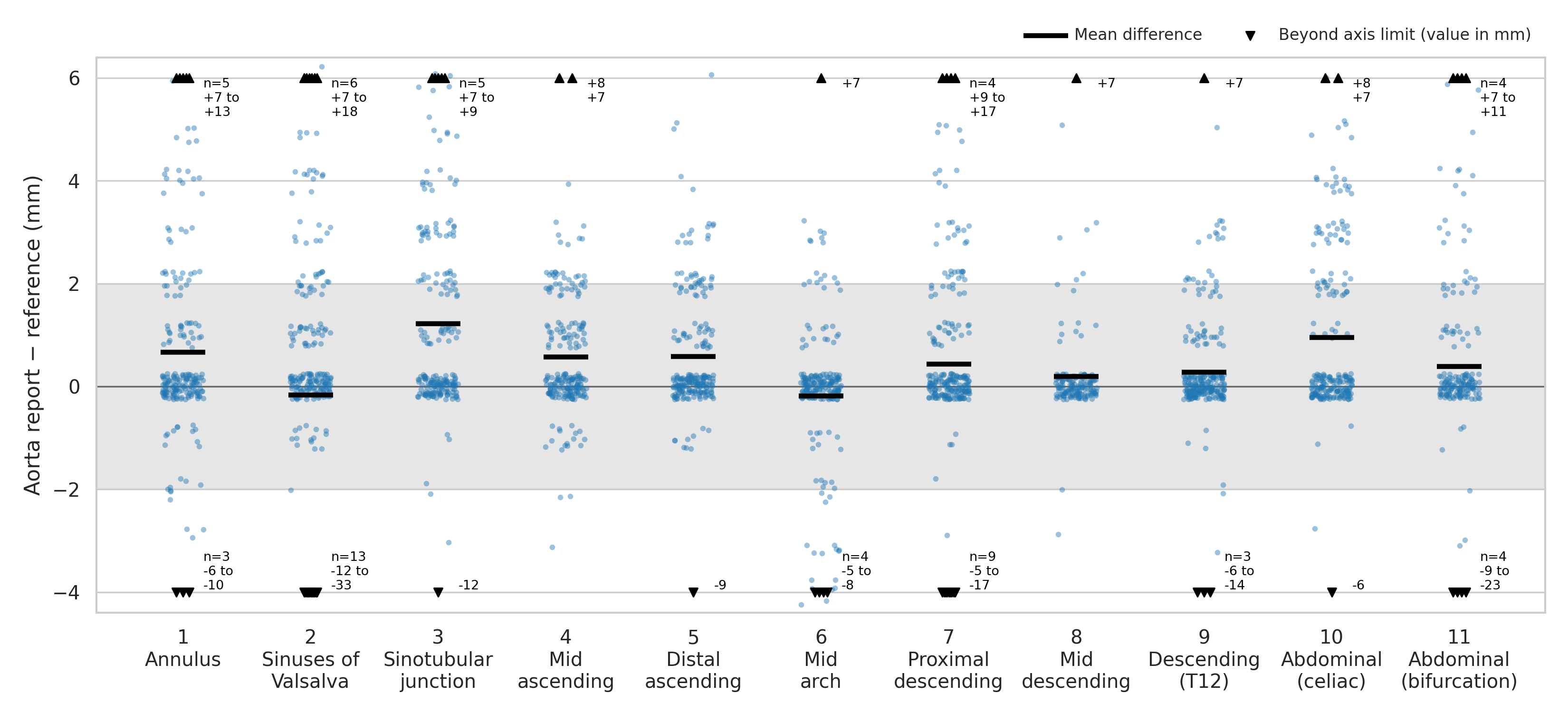}%
  }{%
    \fbox{\parbox{0.92\textwidth}{\centering Missing file: \texttt{imgs/landmark\_differences.png}}}%
  }
  \caption{Difference between the maximum diameter of the aorta report and the reference at each landmark. Each dot is one examination, and black bars are the mean difference. Positive values mean the aorta report measured larger. The gray band marks differences of 2\,mm or less. Both diameters were available in whole millimeters, so a small vertical jitter ($\pm$0.25\,mm) was added for display. The y-axis is limited to the 2nd to 98th percentile of all differences, widened by 1\,mm ($-4$ to 6\,mm). The 70 of 2355 differences (3.0\%) beyond these limits are shown as black triangles at the limit and labeled with their values in millimeters, or with their number (n) and range if there are more than two. Means include all differences.}
  \label{fig:differences}
\end{figure*}

Agreement was best from the mid ascending to the descending aorta at T12, with an MAE of 0.26--0.77\,mm, except at the proximal descending aorta (1.29\,mm) (Table~\ref{tab:landmark}, Fig~\ref{fig:differences}). The aortic root had the largest errors: the MAE was 2.17\,mm at the sinuses of Valsalva, the only landmark above 2\,mm, 1.41\,mm at the sinotubular junction, and 1.20\,mm at the annulus. The sinotubular junction (53 of 220 [24.1\%]) and the celiac level (45 of 219 [20.5\%]) had the highest proportions of differences greater than 2\,mm.

When the two measurements differed, the tool measured larger at 10 of 11 landmarks. This bias was significant at every landmark except the mid aortic arch ($P=.11$) and largest at the sinotubular junction (1.22\,mm) and the celiac level (0.95\,mm). At the sinuses of Valsalva, the mean bias was negative ($-0.16$\,mm) because of a few large underestimations, although the tool measured larger in 83 of the 111 nonequal measurements.

\begin{table*}[t]
  \centering
  \caption{Agreement of the maximum diameter per landmark.}
  \label{tab:landmark}
  \footnotesize
  \begin{tabular}{@{}lrrrrr@{}}
    \toprule
    Landmark & Compared & MAE (mm) & Differences $>$2\,mm & Bias (mm) & Failed measurements \\
    \midrule
    1 Annulus & 217 & 1.20 (0.95, 1.47) & 35/217 (16.1) & 0.66 (0.38, 0.95) & 4/221 (1.8) \\
    2 Sinuses of Valsalva & 227 & 2.17 (1.59, 2.86) & 44/227 (19.4) & $-0.16$ ($-0.90$, 0.50) & 7/234 (3.0) \\
    3 Sinotubular junction & 220 & 1.41 (1.16, 1.67) & 53/220 (24.1) & 1.22 (0.95, 1.50) & 6/226 (2.7) \\
    4 Mid ascending aorta & 223 & 0.77 (0.63, 0.92) & 11/223 (4.9) & 0.57 (0.41, 0.74) & 8/231 (3.5) \\
    5 Distal ascending aorta & 228 & 0.74 (0.59, 0.91) & 18/228 (7.9) & 0.58 (0.41, 0.75) & 0/228 (0) \\
    6 Mid aortic arch & 222 & 0.74 (0.57, 0.94) & 26/222 (11.7) & $-0.19$ ($-0.40$, 0.02) & 0/222 (0) \\
    7 Proximal descending aorta & 232 & 1.29 (0.95, 1.66) & 36/232 (15.5) & 0.43 (0.05, 0.82) & 0/232 (0) \\
    8 Mid descending aorta & 160 & 0.26 (0.13, 0.41) & 6/160 (3.8) & 0.19 (0.07, 0.35) & 0/160 (0) \\
    9 Descending aorta at T12 & 231 & 0.61 (0.44, 0.79) & 16/231 (6.9) & 0.28 (0.07, 0.46) & 0/231 (0) \\
    10 Abdominal aorta, celiac trunk & 219 & 1.05 (0.84, 1.28) & 45/219 (20.5) & 0.95 (0.74, 1.18) & 1/220 (0.5) \\
    11 Abdominal aorta, bifurcation & 176 & 1.23 (0.84, 1.69) & 26/176 (14.8) & 0.39 ($-0.11$, 0.83) & Not counted \\
    \midrule
    All landmarks & 2355 & 1.06 (0.91, 1.22) & 316/2355 (13.4) & 0.45 (0.34, 0.57) & 26/2205 (1.2) \\
    \bottomrule
  \end{tabular}
  \\[2pt]
  \parbox{\textwidth}{\scriptsize Note.---Data in parentheses are 95\% CIs for MAE and bias and percentages for counts. Bias is the mean of the aorta report minus the reference. Positive values mean the aorta report measured larger. Failed measurements are landmarks with a reference diameter but no automated diameter. They were not counted at landmark 11 because the bifurcation is often outside the field of view. The number of compared landmarks varies because not every landmark had a reference diameter. MAE = mean absolute error.}
\end{table*}

\subsection{Overall Agreement}

Across all 2355 landmark measurements, the MAE was 1.06\,mm (95\% CI: 0.91, 1.22). Measurements agreed exactly in 1479 of 2355 (62.8\%), differed by 2\,mm or less in 2039 of 2355 (86.6\%), and differed by 5\,mm or less in 2276 of 2355 (96.6\%). The bias was 0.45\,mm (95\% CI: 0.34, 0.57; $P<.001$), with 95\% limits of agreement of $-4.5$ to 5.4\,mm. Of the 876 nonequal measurements, the tool measured larger in 714 (81.5\%). The examination-level MAE exceeded 2\,mm in 55 of 239 examinations (23.0\% [95\% CI: 18.1, 28.8]).

The tool failed to measure 26 of 2205 landmarks (1.2\%), in 10 of 239 examinations (4.2\% [95\% CI: 2.3, 7.5]). All failures except one were between the annulus and the mid ascending aorta, and no examination failed completely.

The perpendicular diameter agreed slightly better than the maximum diameter: on 2342 paired measurements, the MAE was 0.89 versus 1.05\,mm ($P<.001$), and 89.9\% (95\% CI: 88.2, 91.5) of differences were 2\,mm or less. Its bias was smaller (0.18\,mm [95\% CI: 0.08, 0.28]). The annulus was the exception, with a higher perpendicular MAE (1.44 vs 1.20\,mm).

\subsection{Subgroups}

The examination-level MAE did not differ significantly between examinations with and without any of the six findings (all adjusted $P\geq.70$) (Table~\ref{tab:subgroups}). Examinations with mural thrombus more often had an MAE greater than 2\,mm (11 of 22 [50.0\%] vs 44 of 217 [20.3\%]; adjusted $P=.03$). In these examinations, the MAE was 5.71\,mm (95\% CI: 3.14, 8.43) at the bifurcation, with 10 of 14 differences greater than 2\,mm, and 2.81\,mm at the proximal descending aorta. After aortic graft or stent placement, the MAE at the proximal descending aorta was 3.53\,mm, caused by five of 19 examinations with differences of 8--17\,mm. Examinations with an aortic valve prosthesis more often had at least one failed landmark (three of 10 [30\%] vs seven of 229 [3.1\%]; adjusted $P=.03$). The bias did not differ between examinations with and without any finding (all adjusted $P\geq.31$).

The MAE did not differ significantly across atherosclerosis grades ($P=.95$), nor did the number of failed landmarks ($P=.68$). The MAE did not differ significantly between ECG-gated and nongated examinations (1.11 vs 1.35\,mm; $P=.14$), but the bias was larger without gating (0.97 vs 0.42\,mm; $P=.01$).

\begin{table*}[t]
  \centering
  \caption{Agreement of the maximum diameter by aortic finding, atherosclerosis grade, and ECG gating.}
  \label{tab:subgroups}
  \footnotesize
  \begin{tabular}{@{}lrllrll@{}}
    \toprule
    & & \multicolumn{2}{c}{Examination-level MAE (mm)} & Holm & \multicolumn{2}{c}{Examinations with MAE $>$2\,mm} \\
    \cmidrule(lr){3-4} \cmidrule(lr){6-7}
    Group & $n$ & Present & Absent & $P$ value & Present & Absent \\
    \midrule
    Aneurysm & 114 & 1.21 (0.97, 1.47) & 1.07 & $>.99$ & 32/114 (28.1) & 23/125 (18.4) \\
    Ectasia or dilatation & 89 & 1.11 (0.88, 1.36) & 1.16 & $>.99$ & 17/89 (19.1) & 38/150 (25.3) \\
    Mural thrombus & 22 & 1.65 (1.05, 2.27) & 1.09 & .70 & 11/22 (50.0)\textsuperscript{a} & 44/217 (20.3) \\
    Aortic graft or stent & 22 & 1.42 (0.76, 2.14) & 1.11 & $>.99$ & 7/22 (31.8) & 48/217 (22.1) \\
    Aortic valve prosthesis & 10 & 1.15 (0.08, 2.48) & 1.14 & $>.99$ & 2/10 (20.0) & 53/229 (23.1) \\
    Bicuspid aortic valve & 20 & 1.01 (0.57, 1.46) & 1.15 & $>.99$ & 6/20 (30.0) & 49/219 (22.4) \\
    \midrule
    Atherosclerosis none & 76 & 1.10 (0.84, 1.37) & & .95\textsuperscript{b} & 15/76 (19.7) & \\
    Atherosclerosis mild & 61 & 0.99 (0.76, 1.25) & & & 12/61 (19.7) & \\
    Atherosclerosis moderate & 46 & 1.22 (0.82, 1.67) & & & 11/46 (23.9) & \\
    Atherosclerosis severe & 19 & 1.19 (0.55, 1.89) & & & 6/19 (31.6) & \\
    \midrule
    ECG-gated & 209 & 1.11 (0.94, 1.30) & & .14\textsuperscript{c} & 45/209 (21.5) & \\
    Not ECG-gated & 30 & 1.35 (0.96, 1.74) & & & 10/30 (33.3) & \\
    \bottomrule
  \end{tabular}
  \\[2pt]
  \parbox{\textwidth}{\scriptsize Note.---Examination-level MAE is the mean over the landmarks of an examination. Data in parentheses are 95\% CIs. Counts are numbers of examinations, with percentages in parentheses. $P$ values compare MAE between examinations with and without the finding (Mann-Whitney U test, Holm-adjusted across the six findings). ECG = electrocardiography, MAE = mean absolute error.
  \textsuperscript{a}Higher than without mural thrombus (Fisher exact test, Holm-adjusted $P=.03$). All other findings, adjusted $P\geq.45$.
  \textsuperscript{b}Kruskal-Wallis test across the four grades, not adjusted. The 37 examinations without a grade were not included.
  \textsuperscript{c}Mann-Whitney U test, ECG-gated vs not ECG-gated, not adjusted.}
\end{table*}

\section{Discussion}
\label{sec:discussion}

We developed an open-source tool that measures aortic diameters fully automatically at 11 standardized landmarks from the annulus to the bifurcation and evaluated it in 239 arterial phase CTA examinations. The MAE was 1.06\,mm, 86.6\% of differences were 2\,mm or less, and only 1.2\% of landmarks could not be measured. Agreement was highest from the ascending to the descending aorta and lowest at the aortic root.

The overall error is below the interreader variability reported for CT aortic diameters. Pradella et al reported a human performance variability of up to $\pm$3.4\,mm in repeated measurements under research conditions~\citep{pradella2021}, Hamelink et al found a mean absolute difference of 1.4\,mm between two readers at thoracic landmarks~\citep{hamelink2023}, and 18\% of interreader differences of the infrarenal aorta exceeded 2\,mm in the Troms{\o} study~\citep{singh2003}. Multisociety recommendations advise caution when interpreting diameter changes of 3\,mm or less at ECG-gated CT and 5\,mm or less at nongated CT~\citep{goldstein2015}. Our error is in the range reported for other automated tools against human readers, 1.3--1.6\,mm~\citep{hamelink2023,postiglione2024,rao2026}. These studies used different references and populations, so the values are not directly comparable.

The aortic root was the weakest region, as for every tool that measures it~\citep{pradella2021,graby2023,paulsen2026,lopiccolo2023}. The sinuses of Valsalva have a trilobed cross-section without a single defined diameter, and the measured value depends strongly on the position and orientation of the plane in the short and oblique root. At the sinotubular junction, the tool measured larger in 99 of 105 nonequal measurements, which suggests a systematic difference in plane position or definition rather than random error. In aortic valve prostheses, the root landmarks failed more often, probably because the prosthesis alters the annulus segmentation on which landmarks 1 and 2 depend.

The tool measured slightly larger than the reference at most landmarks. This bias was small (0.45\,mm) compared with the spread of differences and remained despite eroding the mask by one voxel. A likely reason is that the segmentation includes part of the aortic wall, whereas the 2022 ACC/AHA guideline recommends measuring the root and ascending aorta from inner edge to inner edge~\citep{isselbacher2022}. Paulsen et al reported a larger overestimation of up to 2.1\,mm with a TotalSegmentator-based approach~\citep{paulsen2026}. The larger bias without ECG gating (0.97 vs 0.42\,mm) is consistent with motion blurring the vessel boundary.

Accuracy did not decrease with aneurysm, ectasia, prior repair, bicuspid valve, or atherosclerosis, whereas other tools had larger errors in aneurysms, endografts, or after repair~\citep{monti2022,rao2026}. Mural thrombus was the exception: half of these examinations had an MAE greater than 2\,mm, and the error at the bifurcation was 5.7\,mm. One plausible explanation is that the tool and the radiologist placed the boundary differently, at the outer wall or at the patent lumen. The 2022 ACC/AHA guideline recommends measuring the outer margins of segments with wall changes such as mural thrombus~\citep{isselbacher2022}, so these cases need visual review.

Our study had several limitations. First, the reference was created by correcting the automated measurements, not by measuring independently. This can bias the reader toward the automated value and likely overestimates agreement, particularly for small differences. However, it reflects the intended use of the tool, in which a radiologist checks and corrects its output. Second, the study was performed at one center, and nearly all examinations were acquired on one scanner model, so multicenter validation is needed. Third, the tool is designed for any CT image, but only arterial phase CTA was evaluated. Other tools lost accuracy on unenhanced and nongated examinations~\citep{monti2022,graby2023}, so other contrast phases and low-dose CT must be evaluated separately. Fourth, reference diameters were recorded in whole millimeters, which limits the resolution of small differences. Fifth, several subgroups were small and examinations in which the tool had been run on a series that did not cover the entire aorta were excluded.

In conclusion, the aorta report measured aortic diameters at 11 standardized landmarks along the entire aorta with an MAE of about 1\,mm, below published interreader variability. Agreement was lower at the aortic root and in examinations with mural thrombus, where measurements should be reviewed. The tool is available in TotalSegmentator (\url{https://github.com/wasserth/TotalSegmentator}).

\paragraph{Use of AI} Manuscript writing was supported by AI-based language tools. The authors reviewed and edited all content and take full responsibility for the manuscript.

\bibliographystyle{plainnat}


\end{document}